\documentclass[tikz,border=10pt,logo,twocolumn,copyright]{nvidia}

\usepackage{color,xcolor}
\usepackage{epsfig}
\usepackage{graphicx}

\usepackage{adjustbox}
\usepackage{array}
\usepackage{booktabs}
\usepackage{colortbl}
\usepackage{float,wrapfig}
\usepackage{hhline}
\usepackage{multirow}
\usepackage{subcaption}
\usepackage[toc,page]{appendix}
\usepackage[utf8]{inputenc}
\usepackage{pifont}
\usepackage{tikz}
\usetikzlibrary{tikzmark,positioning,arrows.meta,fit}

\usepackage{amsmath,amsfonts,amsthm,amssymb}
\usepackage{bm}
\usepackage{nicefrac}
\usepackage{microtype}
\usepackage{inconsolata}
\usepackage{pifont}
\usepackage{float}

\usepackage{changepage}
\usepackage{extramarks}
\usepackage{fancyhdr}
\usepackage{lastpage}
\usepackage{setspace}
\usepackage{soul}
\usepackage{xspace}
\usepackage{indentfirst}
\usepackage{paralist}
\usepackage{booktabs,arydshln}
\usepackage{makecell}

\usepackage[breaklinks=true,colorlinks,allcolors=blue,bookmarks=false]{hyperref}
\usepackage{url}

\usepackage{algorithm,algorithmic}
\usepackage{enumerate}
\usepackage{lipsum}
\usepackage{tikz}
\usetikzlibrary{tikzmark}

\newcolumntype{L}[1]{>{\raggedright\let\newline\\\arraybackslash\hspace{0pt}}m{#1}}
\newcolumntype{C}[1]{>{\centering\let\newline\\\arraybackslash\hspace{0pt}}m{#1}}
\newcolumntype{R}[1]{>{\raggedleft\let\newline\\\arraybackslash\hspace{0pt}}m{#1}}

\newcommand{\ignorethis}[1]{}

\makeatletter
\DeclareRobustCommand\onedot{\futurelet\@let@token\@onedot}
\def\@onedot{\ifx\@let@token.\else.\null\fi\xspace}

\makeatother

\makeatletter
\def\adl@drawiv#1#2#3{%
        \hskip.5\tabcolsep
        \xleaders#3{#2.5\@tempdimb #1{1}#2.5\@tempdimb}%
                #2\z@ plus1fil minus1fil\relax
        \hskip.5\tabcolsep}
\newcommand{\cdashlinelr}[1]{%
  \noalign{\vskip\aboverulesep
           \global\let\@dashdrawstore\adl@draw
           \global\let\adl@draw\adl@drawiv}
  \cdashline{#1}
  \noalign{\global\let\adl@draw\@dashdrawstore
           \vskip\belowrulesep}}
\makeatother

\definecolor{nvcolor}{HTML}{76b900}
\definecolor{mydarkblue}{rgb}{0,0.08,1}
\definecolor{mydarkgreen}{rgb}{0.02,0.6,0.02}
\definecolor{mydarkred}{rgb}{0.8,0.02,0.02}
\definecolor{mydarkorange}{rgb}{0.40,0.2,0.02}
\definecolor{mypurple}{RGB}{111,0,255}
\definecolor{myred}{rgb}{1.0,0.0,0.0}
\definecolor{mygold}{rgb}{0.75,0.6,0.12}
\definecolor{mydarkgray}{rgb}{0.66, 0.66, 0.66}

\definecolor{darkblue}{rgb}{0,0.08,1}
\definecolor{darkgreen}{rgb}{0.02,0.6,0.02}
\definecolor{darkred}{rgb}{0.8,0.02,0.02}
\definecolor{darkorange}{rgb}{0.40,0.2,0.02}
\definecolor{darkpurple}{RGB}{111,0,255}

\newif\ifdraft
\drafttrue 

\begin{document}

\title{G-SHARP: Gaussian Surgical Hardware Accelerated Real-time Pipeline}
\author{\centering Vishwesh Nath\textsuperscript{1,$\dag$} Javier G. Tejero\textsuperscript{1} Aravind S. Kumar\textsuperscript{1} Ruilong Li \textsuperscript{1} 

 \centering Filippo Filicori \textsuperscript{2} 
 Mahdi Azizian\textsuperscript{1}  Sean D. Huver\textsuperscript{1} \\~\\
\textsuperscript{1}NVIDIA \quad \\
\textsuperscript{2}Northwell Health \quad \\
\textsuperscript{$\dag$}Corresponding Author}

\begin{abstract}

\textbf{Abstract:} \hspace{2pt}
We propose G-SHARP, a commercially compatible, real-time surgical scene reconstruction framework designed for minimally invasive procedures that require fast and accurate 3D modeling of deformable tissue. While recent Gaussian splatting approaches have advanced real-time endoscopic reconstruction, existing implementations often depend on non-commercial derivatives, limiting deployability. G-SHARP overcomes these constraints by being the first surgical pipeline built natively on the GSplat (Apache-2.0) differentiable Gaussian rasterizer, enabling principled deformation modeling, robust occlusion handling, and high-fidelity reconstructions on the EndoNeRF pulling benchmark. Our results demonstrate state-of-the-art reconstruction quality with strong speed–accuracy trade-offs suitable for intra-operative use. Finally, we provide a Holoscan SDK application that deploys G-SHARP on NVIDIA IGX Orin and Thor edge hardware, enabling real-time surgical visualization in practical operating-room settings.

\vspace{5pt}

\textbf{Links:} \hspace{2pt} \href{https://github.com/nvidia-holoscan/holohub/tree/main/applications/surgical_scene_recon}{Code} | \href{https://github.com/nvidia-holoscan/holohub/tree/main}{Holohub}

\vspace{10pt}

\end{abstract}

\maketitle

\section{Introduction}

Minimally invasive surgery demands fast, accurate 3D reconstructions of deformable soft tissue from monocular or stereo endoscopy. While NeRF-style methods (e.g., EndoNeRF) established the benchmark on the EndoNeRF dataset’s pulling and cutting sequences, their training and inference cost limits intra-operative use \cite{wang2022endonerf,wang2024endonerf_journal}. Recent work pivots to Gaussian splatting for real-time rendering with explicit primitives, showing strong results on EndoNeRF benchmarks. Representative methods include EndoGS, EndoGaussian, Endo-4DGS, Deform3DGS, LGS, HFGS, SDFPlane, DESR, ST-Endo4DGS, T-GS, Rational-Wavelet 4DGS, SAGS, EndoWave, EndoSparse, the ACCV GS reconstruction, and EndoPlanar \cite{zhu2024endogs,liu2024endogaussian,huang2024endo4dgs,yang2024deform3dgs,liu2024lgs,zhao2024hfgs,li2024sdfplane,fu2024desr,li2024stendo4dgs,saha2025tgs,li2025rw4dgs,zhang2025sags,wu2025endowave,liu2024endosparse,minh2024accv,endoplanar2025}. However, most implementations rely on kernels or forks derived from the original 3DGS ~\cite{kerbl2023gsplatting} method, which is typically non-commercially licensed or encumbered by third-party components, thus serving out motivation for creating a commercially usable and open-source framework.

To address this limitation, we propose G-SHARP (Gaussian Surgical Holoscan Accelerated Real-time Pipeline), a fully open-source and commercially compatible surgical reconstruction framework. G-SHARP is, to the best of our knowledge, the first surgical reconstruction pipeline that uses the GSplat \cite{ye2024gsplat} (Apache-2.0) backend—an actively maintained, PyTorch-compatible differentiable Gaussian rasterizer from the Nerfstudio project. GSplat couples a simple Python API with highly optimized CUDA kernels, providing a modular and efficient foundation for Gaussian Splatting research. It incorporates engineering improvements for speed, memory usage, and convergence stability over early implementations~\cite{kerbl2023gsplatting}, and offers native PyTorch fallbacks, example-driven documentation, and comprehensive tests for its CUDA operators. By building on GSplat, we inherit a stable rendering backend with modern features such as multi-view batching and memory-efficient rasterization \cite{ye2024gsplat,gsplat_github,gsplat_docs}, while benefiting from its permissive licensing.

On the EndoNeRF pulling sequence \cite{wang2022endonerf,wang2024endonerf_journal}, we demonstrate that a GSplat-native design enables principled deformation modeling (e.g., per-Gaussian constraints, motion decoders) and robust handling of tool–tissue occlusions, achieving state-of-the-art image reconstruction and speed–quality trade-offs while simplifying downstream integration. 

From a clinical perspective accurately modeling tissue deformability has long represented a central bottleneck in the development of high-fidelity surgical simulation and intra-operative (Augmented Reality) AR systems. A major distinction between progress in surgical guidance and other physical-world AI domains—most notably autonomous driving—is the fundamental difference in environmental dynamics: whereas autonomous navigation largely involves interactions with rigid, predictable structures, surgical environments are dominated by highly deformable, anisotropic, and patient-specific soft tissues. Current computational frameworks have struggled to capture these nonlinear tissue behaviors with sufficient realism and temporal responsiveness, limiting the reliability of intra-operative decision-support tools.

Achieving real-time, physiologically plausible simulation of soft-tissue deformation remains a critical milestone for advancing both preoperative training and intra-operative navigation. Resolving tool–tissue interactions under conditions of dynamic occlusion is essential for AR-guided surgery. The inability of many existing AR systems to correctly handle occlusion—especially the occlusion of projected overlays—significantly degrades depth perception, anatomical fidelity, and the surgeon’s situational awareness around target structures.

In summary, our contributions are: (1) a GSplat-backed, commercially compatible surgical gaussian splatting pipeline, (2) a deformation and masking scheme tailored to EndoNeRF-Pulling, (3) state-of-the-art performance with strong real-time characteristics on standard endoscopic benchmarks and (4) real-time holoscan SDK application that allows the utility of gaussian splats for surgery on Nvidia edge hardware (IGX Orin, Thor).
\section{Related Work}

\paragraph{EndoNeRF and EndoNeRF–Pulling.}
EndoNeRF introduced dual-field NeRFs for deformable surgical scenes and established the widely adopted \emph{pulling} and \emph{cutting} benchmarks for dynamic endoscopy \cite{wang2022endonerf,wang2024endonerf_journal}. Following this, a large body of endoscopy-focused 3D Gaussian splatting methods specifically evaluates performance on EndoNeRF–Pulling. EndoGS applies deformable Gaussians with depth and mask supervision \cite{zhu2024endogs}; EndoGaussian emphasizes fast initialization, spatio-temporal Gaussian tracking, and real-time performance \cite{liu2024endogaussian}; Endo-4DGS incorporates depth-prior initialization and confidence-guided learning with explicit Pulling/Cutting tables \cite{huang2024endo4dgs}. Deform3DGS introduces flexible per-Gaussian deformation bases \cite{yang2024deform3dgs}, while LGS proposes lightweight pruning and attribute condensation that significantly reduce Gaussian count \cite{liu2024lgs}. SDFPlane integrates explicit surface modeling with Gaussian splats and reports both qualitative and quantitative results on the Pulling sequence \cite{li2024sdfplane}. DESR evaluates dynamic surgical reconstruction with Gaussian splatting on EndoNeRF Pulling/Cutting within a real-time SPIE setting \cite{fu2024desr}. ST-Endo4DGS provides an unbiased 4DGS formulation for dynamic endoscopic scenes \cite{li2024stendo4dgs}. T-GS extends Gaussian splatting with comprehensive deformation modeling and reports Pulling/Cutting results \cite{saha2025tgs}. Recent higher-frequency or alias-free variants, including Rational-Wavelet 4DGS and SAGS, further refine dynamic fidelity and provide Pulling ablations \cite{li2025rw4dgs,zhang2025sags}. EndoWave introduces rational-wavelet regularization and evaluates on EndoNeRF sequences \cite{wu2025endowave}. Additional work such as EndoSparse and the ACCV GS reconstruction also benchmark on EndoNeRF–Pulling with sparse-view or sequential inputs \cite{liu2024endosparse,minh2024accv}. Planar-based modeling has also emerged via EndoPlanar, which validates its Gaussian-based planar deformation on the Pulling sequence \cite{endoplanar2025}.

\paragraph{Surgical-specific 3DGS design.}
Several works refine Gaussian splatting specifically for surgery through improved geometry alignment, motion consistency, or instrument reasoning. SurgicalGaussian introduces deformable Gaussians combined with tool-aware masking and per-Gaussian motion decoding \cite{xie2024surgicalgaussian}. Surgical Gaussian Surfels extend this idea with surface-aligned, elliptical splats and fast deformation MLPs that better capture tissue topology \cite{sunmola2025sgs}. Both methods benchmark on EndoNeRF–Pulling and highlight the importance of handling tool–tissue interactions and occlusions in surgical reconstruction.

\paragraph{GSplat in the context of differentiable splatting.}
Beyond licensing differences, the choice of rasterization backend has practical consequences for accuracy, stability, and extensibility in surgical reconstruction. The original 3DGS ~\cite{kerbl2023gsplatting} and 4DGS implementations rely on tightly coupled C++/CUDA kernels, which are optimized for fixed splat formulations but more challenging to modify for surgical requirements such as tool-aware occlusion reasoning, per-Gaussian deformation fields, or covariance constraints. Several endoscopy-focused derivatives (e.g., EndoGS, Deform3DGS, LGS) introduce local patches to these kernels, but each creates a forked ecosystem with duplicated or incompatible rendering paths \cite{zhu2024endogs,liu2024lgs,yang2024deform3dgs}. Alternative differentiable renderers such as NVDiffRast, Kaolin, PyTorch3D, or PlenOctrees are well-suited to mesh or voxel primitives but cannot efficiently represent thousands of anisotropic Gaussians with view-dependent covariance, which limits their utility for dynamic soft-tissue modeling.

In contrast, \texttt{gsplat} provides a modular, general-purpose rasterization pipeline built expressly for Gaussian primitives, with support for fast batched rendering, fused forward/backward passes, efficient memory management, and careful numerical designs\cite{ye2024gsplat}. These capabilities are particularly important in surgical settings where Gaussian counts and shapes change rapidly due to deformation, tissue stretch, or instrument occlusion.
Unlike the monolithic 3DGS kernels, \texttt{gsplat} exposes a Python-first API that simplifies integration with deformation models, motion decoders, and per-frame geometric constraints—components heavily used in pipelines such as SurgicalGaussian, SGS, and 4DGS-based variants \cite{xie2024surgicalgaussian,sunmola2025sgs,huang2024endo4dgs}. Prior surgical systems have also reported difficulties extending the original 3DGS implementation to dynamic scenes, including memory growth, covariance instability, and fragile optimization behavior \cite{yang2024deform3dgs,liu2024lgs,xie2024surgicalgaussian}.
\texttt{gsplat}'s combination of fused differentiable kernels, optimized CUDA layouts, and numerically stable gradient computation \cite{ye2024gsplat,gsplat_github,gsplat_docs} therefore provides a more extensible and maintainable foundation for modeling deformable, instrument-rich endoscopic environments—all within a permissive Apache-2.0 license.

\begin{table*}[t]
\centering
\small
\begin{tabular}{l p{10.6cm}}
\hline
Method & One-line description \\
\hline
EndoGS~\cite{zhu2024endogs} & Early deformable Gaussian splatting for endoscopy using depth and mask supervision; reports Pulling/Cutting performance. \\

EndoGaussian~\cite{liu2024endogaussian} & Real-time 3DGS pipeline with holistic Gaussian initialization and spatio-temporal tracking \\

Endo-4DGS~\cite{huang2024endo4dgs} & 4D Gaussian Splatting with depth-prior initialization and confidence-guided updates\\ 

Deform3DGS~\cite{yang2024deform3dgs} & Flexible per-Gaussian deformation bases with dense initialization\\

LGS~\cite{liu2024lgs} & Lightweight 4DGS with Gaussian pruning and deformation-aware attribute reduction\\

HFGS~\cite{zhao2024hfgs} & High-frequency spatial–temporal reconstruction using optical-flow priors\\

SDFPlane~\cite{li2024sdfplane} & Combines explicit SDF planes with Gaussian splats for deformable tissue surfaces\\

DESR~\cite{fu2024desr} & SPIE framework for dynamic endoscopic reconstruction with GS\\

ST-Endo4DGS~\cite{li2024stendo4dgs} & Unbiased spatio-temporal 4DGS formulation with improved regularization\\

T-GS~\cite{saha2025tgs} & MICCAI’25 method for comprehensive dynamic reconstruction using GS\\

Rational-Wavelet 4DGS~\cite{li2025rw4dgs} & Wavelet-regularized 4D Gaussian Splatting with EnodNerf Pulling ablation studies highlighting improved detail recovery. \\

SAGS~\cite{zhang2025sags} & Self-adaptive alias-free Gaussian splatting reducing temporal flicker and splat artifacts\\

EndoWave~\cite{wu2025endowave} & Rational-wavelet dynamic modeling for 4DGS \\

SurgicalGaussian~\cite{xie2024surgicalgaussian} & Deformable 3D Gaussians with tool masking and motion decoders\\

Surgical Gaussian Surfels~\cite{sunmola2025sgs} & Surface-aligned surfel-Gaussians with fused deformation networks\\

EndoSparse~\cite{liu2024endosparse} & Sparse-view Gaussian splatting reconstruction for endoscopy\\

ACCV-GS (Consecutive Frames)~\cite{minh2024accv} & GS-based reconstruction from frame-to-frame correspondences\\

EndoPlanar~\cite{endoplanar2025} & Planar-based deformable GS model for soft tissue motion\\
\hline
\end{tabular}
\caption{Approaches that report evaluations on the EndoNeRF \emph{Pulling} dataset.}
\end{table*}
\section{Method \& Datasets}

\begin{figure}[!b]
    \centering
    \includegraphics[width=\linewidth]{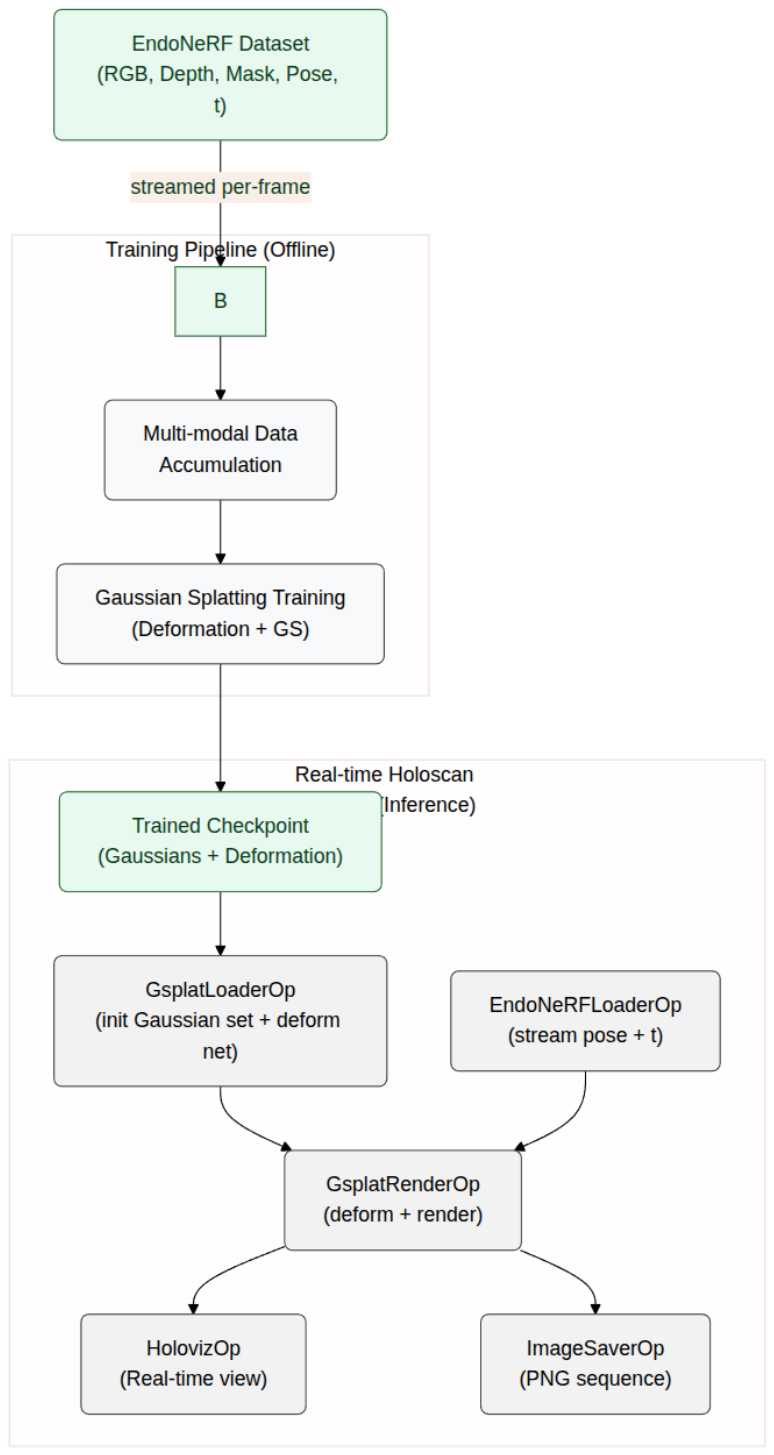}
    \caption{\textbf{Figure:} Training (top) and rendering (bottom) pipelines for temporal Gaussian Splatting of surgical scenes. 
The training phase uses RGB, depth, and segmentation masks from all frames to learn 3D Gaussians and a temporal deformation network. 
The real-time rendering pipeline implements a Holoscan streaming architecture with custom operators: \texttt{EndoNeRFLoaderOp} streams camera poses and timestamps frame-by-frame, \texttt{GsplatLoaderOp} loads the trained checkpoint once at startup, \texttt{GsplatRenderOp} applies temporal deformation and performs differentiable rasterization, \texttt{HolovizOp} provides GPU-accelerated visualization, and \texttt{ImageSaverOp} logs outputs to disk.}
    \label{fig:placeholder}
\end{figure}

\subsection{Dataset}

The EndoNeRF dataset~\cite{wang2022endonerf} provides high-quality stereo endoscopic video sequences captured during minimally invasive surgical procedures, designed for neural rendering and 3D reconstruction tasks. The \textit{pulling soft tissues} sequence captures a surgical scene where instruments interact with and deform soft tissue over time, providing realistic examples of non-rigid tissue dynamics. Each frame in the dataset includes RGB images at \(640 \times 512\) resolution, stereo-derived depth maps, and binary segmentation masks that distinguish surgical tools from anatomical tissue. The sequence consists of 63 temporally ordered frames with calibrated camera intrinsics and extrinsics, enabling accurate 3D reconstruction and temporal modeling of tissue deformation. This dataset is particularly well-suited for training spatiotemporal Gaussian Splatting models, as it provides the multi-modal supervision (color, geometry, and tool segmentation) necessary for learning tool-free tissue reconstruction with temporal consistency.

\subsection{Method Overview}

Our two-stage Gaussian splatting framework addresses three critical challenges in surgical scene reconstruction: narrow field-of-view endoscopic cameras, surgical tool occlusion, and dynamic tissue deformation. The method operates in dual modes: tissue-only reconstruction (tools removed) for surgical planning, and full-scene reconstruction (tissue + tools) for video analysis.

The pipeline begins with multi-frame point cloud initialization, accumulating depth and color across all frames rather than using single-frame structure-from-motion. For tissue-only mode, tool masks exclude instrument regions during accumulation, ensuring 3-5$\times$ more tissue points and complete coverage of chronically occluded areas. For full-scene mode, all regions are included without masking. This initialization strategy provides superior geometric foundation compared to traditional single-frame approaches.

During training, the system employs a novel masked supervision strategy. In tissue-only mode, tool masks are applied to both ground truth and rendered images before loss computation, ensuring gradient flow only through tissue regions. This prevents the network from learning surgical instrument appearance. An ``invisible mask'' is created from the union of all tool positions across frames, identifying chronically occluded regions. Targeted total variation (TV) regularization is applied exclusively to these regions, encouraging smooth, plausible tissue interpolation where direct supervision is unavailable.

The deformation network handles temporal dynamics, modeling tissue motion and tool movement. Crucially, deformation is applied only during the fine training stage and only to tissue-region Gaussians, tracked via a per-Gaussian deformation table. This selective approach prevents tool-region geometry from interfering with tissue reconstruction while maintaining temporal consistency.

\subsection{Network Architecture}

Our 3D representation extends standard Gaussian splatting with temporal deformation capabilities. Each Gaussian primitive is parameterized by position $\boldsymbol{\mu} \in \mathbb{R}^3$, covariance (via scales $\boldsymbol{s} \in \mathbb{R}^3$ and rotation quaternion $\boldsymbol{q} \in \mathbb{R}^4$), opacity $\alpha \in [0,1]$, and spherical harmonic (SH) coefficients $\boldsymbol{c} \in \mathbb{R}^{(l+1)^2 \times 3}$ for view-dependent color up to degree $l=3$.

The deformation network consists of two components: a HexPlane-based spatial-temporal feature grid and a multi-layer perceptron (MLP) decoder. The HexPlane representation factorizes the 4D space-time volume into six 2D feature planes: three spatial planes (XY, XZ, YZ) and three space-time planes (XT, YT, ZT), with resolution $64^3 \times 100$ for spatial and temporal dimensions respectively. Multi-resolution grids at scales $[1, 2, 4, 8]$ capture features at different granularities.

For a Gaussian at canonical position $\boldsymbol{\mu}$ and time $t$, the deformation network computes feature embeddings from the HexPlane grids via bilinear interpolation. The MLP decoder (8 layers, 256-dimensional hidden units) processes these features to predict deformation offsets: $\Delta\boldsymbol{\mu}$ (position), $\Delta\boldsymbol{s}$ (scale), $\Delta\boldsymbol{q}$ (rotation), and $\Delta\alpha$ (opacity). The final deformed parameters are: $\boldsymbol{\mu}' = \boldsymbol{\mu} + \Delta\boldsymbol{\mu}$, and similarly for other attributes.

Rendering follows the standard splatting pipeline: Gaussians are projected to image space, sorted by depth, and alpha-composited front-to-back. For endoscopic cameras, we use pinhole projection with camera intrinsics $\mathbf{K}$ extracted from the EndoNeRF format. Both RGB and expected depth are rendered via weighted blending.

\subsection{Training Configuration}

Training proceeds in two stages. The \textbf{coarse stage} (200 iterations, $\sim$30 seconds) initializes geometry using a fixed camera viewpoint with active densification but no deformation. The \textbf{fine stage} (1500 iterations, $\sim$1 min 30 secs) refines appearance and deformation using random camera sampling from training frames.

The composite loss function is:
\begin{equation}
\mathcal{L} = \mathcal{L}_{\text{L1}} + \lambda_{\text{depth}} \mathcal{L}_{\text{depth}} + \lambda_{\text{SSIM}} \mathcal{L}_{\text{SSIM}} + \lambda_{\text{TV}} \mathcal{L}_{\text{TV}} + \mathcal{L}_{\text{deform}}
\end{equation}

Loss weights are: $\lambda_{\text{depth}} = 0.001$ (minimal depth constraint, RGB-focused), $\lambda_{\text{SSIM}} = 0.2$ (structural similarity), $\lambda_{\text{TV}} = 0.03$ (smoothness). For binocular depth, $\mathcal{L}_{\text{depth}}$ computes L1 loss in disparity space; for monocular depth, it uses Pearson correlation. The TV loss is applied only to masked regions in tissue-only mode.

Deformation regularization includes: time smoothness weight $0.01$, L1 temporal planes weight $0.01$, and spatial plane TV weight $0.01$, preventing unrealistic deformations and encouraging temporal coherence.

Gaussian densification operates from iteration 500 to 15,000, with refinement every 100 iterations and opacity reset every 3,000 iterations. Densification thresholds decay during fine stage: opacity threshold from $0.05 \to 0.005$, gradient threshold $0.0002$ (constant). Learning rates are: means $1.6 \times 10^{-4} \cdot \text{scene\_scale}$, scales $5 \times 10^{-3}$, opacities $5 \times 10^{-2}$, quaternions $1 \times 10^{-3}$, SH coefficients $2.5 \times 10^{-3}$ (DC) and $1.25 \times 10^{-4}$ (higher orders). Deformation network uses separate learning rates: MLP $1 \times 10^{-5}$ and HexPlane grids $1 \times 10^{-5}$, both scaled by scene extent.

All optimizers use Adam with $\beta_1 = 0.9$, $\beta_2 = 0.999$, and $\epsilon = 10^{-15}$. Images are processed at native resolution ($640 \times 512$ for EndoNeRF). The system achieves PSNR 24-27 dB (full scene) and 17-19 dB (tissue-only, masked comparison) on EndoNeRF datasets.

\subsection{Holoscan Application}

The holoscan application framework can be observed in Figure 1. The key insight is that training consumes all data modalities to learn the 3D representation, while the Holoscan rendering pipeline only requires camera poses and time—the appearance, geometry, and deformation are encoded in the trained checkpoint and streamed through specialized operators for real-time performance.

The real-time Holoscan rendering pipeline requires only camera poses and timestamps, applying learned deformations to generate photorealistic tissue reconstructions. This separation enables efficient deployment where the heavy multi-modal data processing happens offline, while real-time inference operates on minimal pose inputs.

\section{Results}

Table~2 summarizes quantitative performance on the EndoNeRF-Pulling benchmark, comparing reconstruction quality, training efficiency, and real-time capability across representative NeRF- and Gaussian-splatting–based methods. Classical neural-field approaches such as EndoNeRF and EndoSurf achieve PSNRs of 35.43\,dB and 34.91\,dB, respectively, but require 6--7 hours of training per scene and do not reach real-time inference rates. LerPlane-32k improves training efficiency to 8 minutes but sacrifices reconstruction fidelity, dropping to 31.77\,dB. Recent 3DGS approaches—including Endo-4DGS, EndoGaussian, and Deform3DGS—substantially reduce training cost (1--4 minutes) while achieving strong reconstruction accuracy around 37.8--37.9\,dB, all exceeding the 60\,FPS threshold. Surgical Gaussian Surfels attains the highest PSNR (39.06\,dB) but at the cost of a significantly longer 45-minute training time and a non-commercial license constraint. In contrast, \textit{G-SHARP matches state-of-the-art performance (37.98\,dB) with a 2-minute training time, provides real-time rendering above 60\,FPS, and is the only method offering a fully commercial-compatible license}, highlighting its balance of accuracy, efficiency, and deployability.

\begin{table*}[t]
\centering
\small
\begin{tabular}{l c c c c}
\hline
Method & Train time / scene & PSNR (Pulling) & FPS > 60 & License \\
\hline
EndoNeRF~\cite{wang2022endonerf_original} & $\sim$6h & 35.43 & \ding{55} & Commercial \\
EndoSurf~\cite{wang2024endosurf} & $\sim$7h & 34.91 & \ding{55} & MIT \\
LerPlane-32k~\cite{huang2023lerplane} & 8 min & 31.77 & \ding{55} & BSD-3 \\
Endo-4DGS~\cite{huang2024endo4dgs} & 4 min & 37.85 & \checkmark & NC Derivative \\
EndoGaussian~\cite{liu2024endogaussian} & $\le$ 2 min & 37.848 & \checkmark & NC Derivative \\
Deform3DGS~\cite{yang2024deform3dgs} & $\approx$ 1 min & 37.90 & \checkmark & NC Derivative \\
Surgical Gaussian Surfels~\cite{sunmola2025sgs} & 45 min & 39.059 & \checkmark & NC Derivative \\
\rowcolor{gray!10}
G-SHARP (ours) & 2 min & 37.98 & \checkmark & Fully Commercial  \\
\hline
\end{tabular}
\caption{Training time, Pulling-sequence PSNR, and real-time capability (FPS > 60) for methods on Endonerf-pulling dataset}
\end{table*}














\begin{figure}[t]
    \centering
    \includegraphics[width=\linewidth]{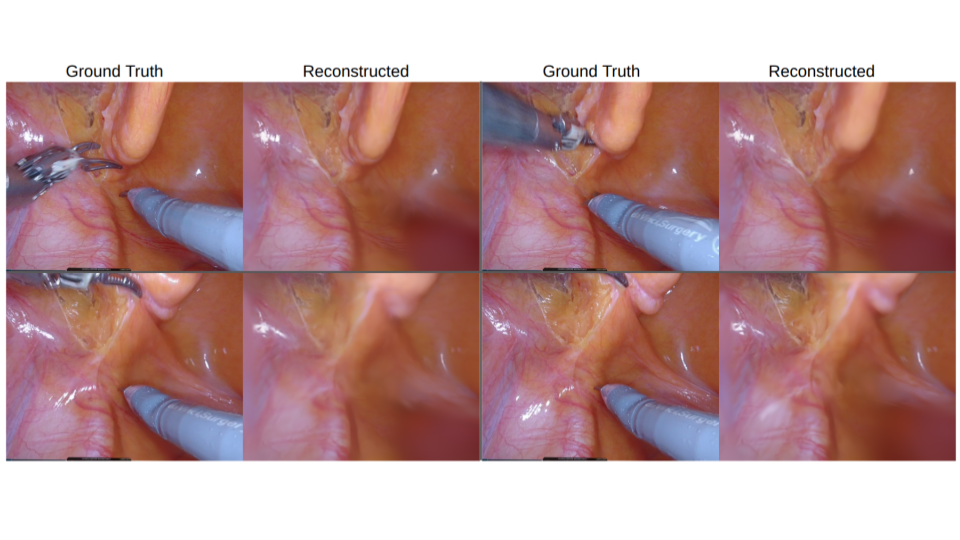}
    \caption{\textbf{Figure:} Ground truth frames with tools are shown on the left and the reconstruction from G-SHARP are shown on the right for different chose frames}
    \label{fig:placeholder}
\end{figure}
\section{Future Direction}

Future extensions of this work include further reducing end-to-end latency by optimizing the GSplat rendering path, enabling sub–10\,ms update rates critical for fully immersive surgical applications. In parallel, integrating the reconstructed surgical scenes into XR/AR environments—such as optical-see-through head-mounted displays or spatial computing platforms—will allow real-time, depth-accurate overlays that enhance intraoperative perception. Another promising direction is coupling the reconstructed 3D surgical fields with vision–language models, enabling contextual scene understanding, autonomous annotation, and intelligent assistance directly within the AR/XR interface. Continued advances in active Gaussian pruning, fast deformation solvers, and streaming-based rendering will further support high-fidelity mixed-reality guidance in dynamic surgical settings.

The convergence of next-generation robotic platforms with integrated haptic force sensing now provides a unique opportunity to bridge these gaps. By combining continuous force tracking with deformation-aware video analysis, future systems will be capable of generating real-time, patient-specific biomechanical models that update synchronously with surgical manipulation. Such capabilities stand to enable a new class of applications, including ultra-realistic simulation environments, robust intraoperative navigation pipelines, and ultimately semi-autonomous or autonomous robotic behaviors grounded in accurate predictions of tissue response.

\textbf{Acknowledgements:} This work was inspired by prior advances in Gaussian splatting and surgical scene reconstruction, particularly the original 3D Gaussian Splatting framework~\cite{kerbl2023gsplatting}, Surgical Gaussian Surfels~\cite{sunmola2025sgs}, and EndoGaussian~\cite{liu2024endogaussian}. These methods motivated our exploration of deformable Gaussian representations and real-time rendering strategies for endoscopic environments. We gratefully acknowledge their foundational contributions to the development of G-SHARP.

{
\small
\bibliographystyle{unsrt}
\bibliography{main}
}

\appendix
\renewcommand{\thesection}{A.\arabic{section}}
\renewcommand{\thetable}{A\arabic{table}}
\renewcommand{\thefigure}{A\arabic{figure}}
\setcounter{section}{0}
\setcounter{table}{0}
\setcounter{figure}{0}

\clearpage

\end{document}